\documentclass{article}

\usepackage[preprint]{neurips_2024}

\usepackage[utf8]{inputenc} 
\usepackage[T1]{fontenc}    
\usepackage{hyperref}       
\usepackage{url}            
\usepackage{booktabs}       
\usepackage{amsfonts}       
\usepackage{nicefrac}       
\usepackage{microtype}      
\usepackage{xcolor}         
\usepackage{graphicx}       

\title{Exposing Assumptions in AI Benchmarks through Cognitive Modelling}

\author{
  Jonathan H. Rystrøm \\
  Oxford Internet Institute \\
  University of Oxford \\
  \texttt{jonathan.rystrom@oii.ox.ac.uk} \\
  \And
  Kenneth C. Enevoldsen \\
  Centre for Humanities Computing \\
  Aarhus University \\
  \texttt{kenneth.enevoldsen@cas.au.dk}
}

\begin{document}

\maketitle

\begin{abstract}
Cultural AI benchmarks often rely on implicit assumptions about measured constructs, leading to vague formulations with poor validity and unclear interrelations. We propose exposing these assumptions using explicit cognitive models formulated as Structural Equation Models. Using cross-lingual alignment transfer as an example, we show how this approach can answer key research questions and identify missing datasets. This framework grounds benchmark construction theoretically and guides dataset development to improve construct measurement. By embracing transparency, we move towards more rigorous, cumulative AI evaluation science, challenging researchers to critically examine their assessment foundations.
\end{abstract}

\paragraph{Introduction}
Following the rapid advancements in Generative AI, marked by the release of ChatGPT, benchmarks for such models have proliferated, encompassing a range of concepts including `reasoning' \citep{sprague_musr_2024} and `cultural alignment' \citep{wang_seaeval_2023}. However, combining insights from these diverse benchmarks remains challenging due to unclear measurement targets and methodologies \citep{jacobs_measurement_2021, raji_ai_2021}.

Psychometrics, experienced in measuring complex constructs, offers valuable approaches for addressing these challenges \citep{jacobs_measurement_2021,hughes_psychometric_2018}. Previous work has applied psychometric methods to LLM benchmarking, defining intelligence as skill acquisition efficiency \citep{chollet_measure_2019} and reducing sample sizes for LLM evaluation \citep{polo2024tinybenchmarks}. However, most explicit psychometric modelling has focused on applying human-oriented psychometric batteries to generative models \citep{hagendorff_machine_2024,jiang_evaluating_2024,serapio-garcia_personality_2023}.

We propose extending psychometrics-inspired approaches to elucidate Large Language Model (LLM) `traits', encompassing capabilities and attributes with no clear `better' or `worse' \citep[e.g., political stances;][]{rottger_political_2024}. Our contribution lies in explicitly using cognitive models \citep{wilson2019ten}, operationalized through Structural Equation Modeling \citep[SEM;][]{suhr_basics_2006}, to expose assumptions in how test batteries relate to theoretical constructs. This approach enables rigorous aggregation of multiple datasets, identifies gaps in existing benchmarks, and develops a sounder theoretical foundation for understanding LLM traits \citep{shanahan_talking_2022}.

To illustrate, we (re)construct the concept of `cultural alignment' using SEM, focusing on cross-lingual alignment transfer. This example demonstrates how our framework improves construct validity, highlights missing datasets, and provides clear testable hypotheses. By offering a meta-evaluation perspective, our approach contributes to the workshop's goal of evaluating AI evaluations, providing a robust framework for assessing traits and their relationships across model evaluations.

\paragraph{(Re)constructing Crosslingual Alignment Transfer}

\begin{figure}
    \centering
    \includegraphics[width=0.9\linewidth]{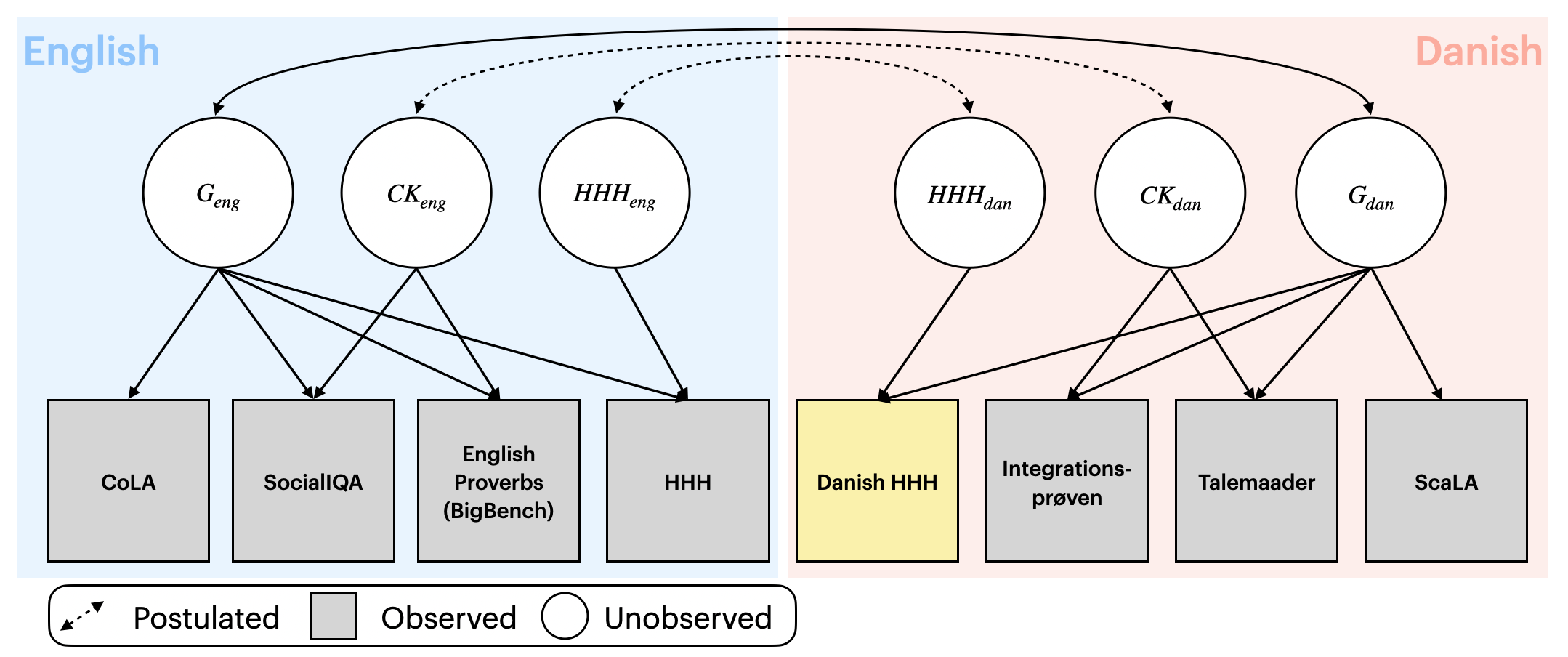}
    \caption{Simplified Structural Equation Model: Cross-lingual Alignment Transfer.
Latent variables (circles): Language ability ($G$), Cultural knowledge ($CK$),
Alignment ($HHH$) for English and Danish. Observable variables (rectangles):
Language-specific tasks (e.g., CoLA, SocialIQA). Arrows: single-headed for hypothesised
relationships, bidirectional for covariance between latent variables.}
    \label{fig:main-concept-model}
\end{figure}

Building on the need for improved evaluation methodologies outlined in the introduction, we now turn to a specific case study in cross-lingual alignment transfer. The concept of `alignment' in LLMs is predominantly studied using English data \citep{kirk_prism_2024}, implicitly assuming cross-lingual transferability. However, recent empirical evidence challenges this assumption \citep{masoud_cultural_2023,tao_cultural_2023,cao_assessing_2023}. Our Structural Equation Model (SEM) in Fig. \ref{fig:main-concept-model} makes this assumption explicit, allowing us to rigorously measure the effect of alignment transfer.

Our model uses unobservable latent factors (circles) for underlying concepts and observable variables (rectangles) for specific tests following \citep{suhr_basics_2006}. We explore relationships between language ability ($G_{eng}$, $G_{dan}$), cultural knowledge ($CK_{eng}$, $CK_{dan}$), and alignment ($HHH_{eng}$, $HHH_{dan}$) for both English and Danish. Arrows indicate potential relationships or influences between these factors. A full toy example of an SEM can be found in Appendix \ref{app:simplified-sem} and explanations of datasets in \ref{app:benchmark}.

This explicit model elucidates several key aspects of alignment transfer:

\begin{enumerate}
    \item \textbf{Design direction:} To assess cross-lingual transfer in alignment, we require a dataset that `loads' on Danish HHH. The model makes this requirement explicit, highlighting an important direction for dataset development.
    \item \textbf{Testable hypotheses:} Our model enables rigorous testing of hypotheses using benchmark data from multiple models, such as whether HHH-alignment transfers across languages and whether there exists an underlying language ability factor.
    \item \textbf{Improved construct validity:} By connecting multiple benchmarks to underlying constructs, we can assess how well these tests measure what they're intended to measure. This allows us to investigate whether different traits converge or if a trait needs to be decomposed into multiple components.
\end{enumerate}

These benefits extend beyond alignment transfer; similar models could deepen our understanding of other phenomena, such as relationships between different biases \citep{kirk_bias_2021}.

Importantly, cognitive modelling cannot define the meaning of constructs on its own. While it can reveal relationships between latent variables, it cannot explain what these traits actually mean. SEM is no replacement for understanding the discussions and defining the concepts \citep{bender_climbing_2020,shanahan_talking_2022}. For example, in our model, `cultural knowledge' as measured by benchmarks could actually represent `memorization' \citep{prashanth_recite_2024} or something else entirely. Careful analysis is needed to distinguish these possibilities. However, cognitive modeling does help by making assumptions explicit. We discuss other issues in the `Limitations' section.

By adopting this psychometric-inspired approach, we propose a path toward a more rigorous, theoretically grounded understanding of LLM capabilities and traits. This framework not only guides benchmark development but also facilitates cumulative science in LLM evaluation \citep{kuhn_structure_1997}. Ultimately, it promotes a more comprehensive and nuanced assessment of Generative AI systems.

\paragraph{Limitations} \label{limitations}
As mentioned in the previous section, a key challenge with our approach is avoiding the `Formalism Trap' \citep{selbst_fairness_2019}. In machine learning, it's common to transform structural problems into computational ones, masking injustice behind algorithmic opacity \citep{burrell_how_2016,malik_hierarchy_2020,buolamwini_gender_2018}. We must ensure cognitive modelling doesn't become yet another blind justification for technical research \citep{birhane_values_2022}. However, when LLM researchers employ mathematical formalism to operationalise benchmarks or make claims about model performance, our framework aims to make their assumptions as explicit as possible. We believe cognitive modelling is a valuable tool for this purpose.

Mathematical formalism can render discussions inaccessible to many communities, particularly those most affected by AI systems \citep{birhane_power_2022}. A key aspect of our approach is using graphical representations instead of hiding assumptions behind equations. For instance, examining Fig \ref{fig:main-concept-model} should get the reader to raise multiple concerns: Is `alignment' and `cultural knowledge' really independent \citep[Likely not;][]{kirk_prism_2024}; aren't there many more factors? etc. When done well, these visual representations can make AI assumption discussions more accessible. This aligns with \citet{rudin_stop_2019}'s call for interpretable machine learning; a simpler model is easier to critique than an opaque one.

Existing research inspired by psychometrics which tests human-like constructs such as 'personality' \citep{jiang_evaluating_2024,serapio-garcia_personality_2023} using tests designed for humans \citep{hagendorff_machine_2024}. This risks anthropomorphising LLMs and importing the problematic history of psychometrics \citep{pilgrim_eugenic_2008,bender_dangers_2021}. While we've likely anthropomorphised LLMs in this paper (as it's tempting to do so), we believe cognitive modelling provides tools to formally construct an appropriate vocabulary that more accurately reflects the `traits' of interest \citep{shanahan_talking_2022} rather than potentially flawed human constructs \citep{buzsaki2019brain}.

Finally, in this work, we choose SEMs to formalise our cognitive models. However, they could similarly have been formalised using alternative methods such as hierarchical Bayesian models \citep{lee_how_2011}. The fundamental idea is to have an explicit model of concepts and how they relate to observable measures \citep{wilson2019ten}. Similarly, in this work, we use aggregated scores as these are often publicly available. However, latent constructs could similarly be estimated pr. sample. This is also noted by  \cite{polo2024tinybenchmarks} which shows that benchmarks can be notably reduced by considering per-sample information.

\bibliographystyle{unsrtnat}
\bibliography{references} 

\begin{thebibliography}{52}
\providecommand{\natexlab}[1]{#1}
\providecommand{\url}[1]{\texttt{#1}}
\expandafter\ifx\csname urlstyle\endcsname\relax
  \providecommand{\doi}[1]{doi: #1}\else
  \providecommand{\doi}{doi: \begingroup \urlstyle{rm}\Url}\fi

\bibitem[Sprague et~al.(2024)Sprague, Ye, Bostrom, Chaudhuri, and Durrett]{sprague_musr_2024}
Zayne Sprague, Xi~Ye, Kaj Bostrom, Swarat Chaudhuri, and Greg Durrett.
\newblock {MuSR}: {Testing} the {Limits} of {Chain}-of-thought with {Multistep} {Soft} {Reasoning}, March 2024.
\newblock URL \url{http://arxiv.org/abs/2310.16049}.
\newblock arXiv:2310.16049 [cs].

\bibitem[Wang et~al.(2023)Wang, Liu, Huang, Jiao, Ding, Aw, and Chen]{wang_seaeval_2023}
Bin Wang, Zhengyuan Liu, Xin Huang, Fangkai Jiao, Yang Ding, AiTi Aw, and Nancy~F. Chen.
\newblock {SeaEval} for {Multilingual} {Foundation} {Models}: {From} {Cross}-{Lingual} {Alignment} to {Cultural} {Reasoning}.
\newblock 2023.
\newblock \doi{10.48550/ARXIV.2309.04766}.
\newblock URL \url{https://arxiv.org/abs/2309.04766}.
\newblock Publisher: arXiv Version Number: 5.

\bibitem[Jacobs and Wallach(2021)]{jacobs_measurement_2021}
Abigail~Z. Jacobs and Hanna Wallach.
\newblock Measurement and {Fairness}.
\newblock In \emph{Proceedings of the 2021 {ACM} {Conference} on {Fairness}, {Accountability}, and {Transparency}}, {FAccT} '21, pages 375--385, New York, NY, USA, March 2021. Association for Computing Machinery.
\newblock ISBN 978-1-4503-8309-7.
\newblock \doi{10.1145/3442188.3445901}.
\newblock URL \url{https://dl.acm.org/doi/10.1145/3442188.3445901}.

\bibitem[Raji et~al.(2021)Raji, Bender, Paullada, Denton, and Hanna]{raji_ai_2021}
Inioluwa~Deborah Raji, Emily~M. Bender, Amandalynne Paullada, Emily Denton, and Alex Hanna.
\newblock {AI} and the {Everything} in the {Whole} {Wide} {World} {Benchmark}, November 2021.
\newblock URL \url{http://arxiv.org/abs/2111.15366}.
\newblock arXiv:2111.15366 [cs].

\bibitem[Hughes(2018)]{hughes_psychometric_2018}
David~J. Hughes.
\newblock Psychometric validity: {Establishing} the accuracy and appropriateness of psychometric measures.
\newblock In \emph{The {Wiley} handbook of psychometric testing: {A} multidisciplinary reference on survey, scale and test development, {Vols}. 1-2}, pages 751--779. Wiley Blackwell, 2018.
\newblock ISBN 978-1-118-48982-6 978-1-118-48983-3 978-1-118-48970-3 978-1-119-12117-6.
\newblock \doi{10.1002/9781118489772.ch24}.

\bibitem[Chollet(2019)]{chollet_measure_2019}
François Chollet.
\newblock On the {Measure} of {Intelligence}, November 2019.
\newblock URL \url{http://arxiv.org/abs/1911.01547}.
\newblock arXiv:1911.01547 [cs].

\bibitem[Polo et~al.(2024)Polo, Weber, Choshen, Sun, Xu, and Yurochkin]{polo2024tinybenchmarks}
Felipe~Maia Polo, Lucas Weber, Leshem Choshen, Yuekai Sun, Gongjun Xu, and Mikhail Yurochkin.
\newblock tinybenchmarks: evaluating llms with fewer examples.
\newblock \emph{arXiv preprint arXiv:2402.14992}, 2024.

\bibitem[Hagendorff et~al.(2024)Hagendorff, Dasgupta, Binz, Chan, Lampinen, Wang, Akata, and Schulz]{hagendorff_machine_2024}
Thilo Hagendorff, Ishita Dasgupta, Marcel Binz, Stephanie C.~Y. Chan, Andrew Lampinen, Jane~X. Wang, Zeynep Akata, and Eric Schulz.
\newblock Machine {Psychology}, August 2024.
\newblock URL \url{http://arxiv.org/abs/2303.13988}.
\newblock arXiv:2303.13988 [cs].

\bibitem[Jiang et~al.(2024)Jiang, Xu, Zhu, Han, Zhang, and Zhu]{jiang_evaluating_2024}
Guangyuan Jiang, Manjie Xu, Song-Chun Zhu, Wenjuan Han, Chi Zhang, and Yixin Zhu.
\newblock Evaluating and inducing personality in pre-trained language models.
\newblock \emph{Advances in Neural Information Processing Systems}, 36, 2024.
\newblock URL \url{https://proceedings.neurips.cc/paper_files/paper/2023/hash/21f7b745f73ce0d1f9bcea7f40b1388e-Abstract-Conference.html}.

\bibitem[Serapio-García et~al.(2023)Serapio-García, Safdari, Crepy, Sun, Fitz, Romero, Abdulhai, Faust, and Matarić]{serapio-garcia_personality_2023}
Greg Serapio-García, Mustafa Safdari, Clément Crepy, Luning Sun, Stephen Fitz, Peter Romero, Marwa Abdulhai, Aleksandra Faust, and Maja Matarić.
\newblock Personality {Traits} in {Large} {Language} {Models}, September 2023.
\newblock URL \url{http://arxiv.org/abs/2307.00184}.
\newblock arXiv:2307.00184 [cs].

\bibitem[Röttger et~al.(2024)Röttger, Hofmann, Pyatkin, Hinck, Kirk, Schütze, and Hovy]{rottger_political_2024}
Paul Röttger, Valentin Hofmann, Valentina Pyatkin, Musashi Hinck, Hannah~Rose Kirk, Hinrich Schütze, and Dirk Hovy.
\newblock Political {Compass} or {Spinning} {Arrow}? {Towards} {More} {Meaningful} {Evaluations} for {Values} and {Opinions} in {Large} {Language} {Models}, June 2024.
\newblock URL \url{http://arxiv.org/abs/2402.16786}.
\newblock arXiv:2402.16786 [cs].

\bibitem[Wilson and Collins(2019)]{wilson2019ten}
Robert~C Wilson and Anne~GE Collins.
\newblock Ten simple rules for the computational modeling of behavioral data.
\newblock \emph{Elife}, 8:\penalty0 e49547, 2019.

\bibitem[Suhr(2006)]{suhr_basics_2006}
Diana Suhr.
\newblock The basics of structural equation modeling.
\newblock \emph{Presented: Irvine, CA, SAS User Group of the Western Region of the United States (WUSS)}, pages 1--19, 2006.
\newblock URL \url{https://www.lexjansen.com/wuss/2006/tutorials/TUT-Suhr.pdf}.

\bibitem[Shanahan(2022)]{shanahan_talking_2022}
Murray Shanahan.
\newblock Talking {About} {Large} {Language} {Models}, December 2022.
\newblock URL \url{http://arxiv.org/abs/2212.03551}.
\newblock arXiv:2212.03551 [cs].

\bibitem[Kirk et~al.(2024)Kirk, Whitefield, Röttger, Bean, Margatina, Ciro, Mosquera, Bartolo, Williams, He, Vidgen, and Hale]{kirk_prism_2024}
Hannah~Rose Kirk, Alexander Whitefield, Paul Röttger, Andrew Bean, Katerina Margatina, Juan Ciro, Rafael Mosquera, Max Bartolo, Adina Williams, He~He, Bertie Vidgen, and Scott~A. Hale.
\newblock The {PRISM} {Alignment} {Project}: {What} {Participatory}, {Representative} and {Individualised} {Human} {Feedback} {Reveals} {About} the {Subjective} and {Multicultural} {Alignment} of {Large} {Language} {Models}.
\newblock 2024.
\newblock \doi{10.48550/ARXIV.2404.16019}.
\newblock URL \url{https://arxiv.org/abs/2404.16019}.
\newblock Publisher: arXiv Version Number: 1.

\bibitem[Masoud et~al.(2023)Masoud, Liu, Ferianc, Treleaven, and Rodrigues]{masoud_cultural_2023}
Reem~I. Masoud, Ziquan Liu, Martin Ferianc, Philip Treleaven, and Miguel Rodrigues.
\newblock Cultural {Alignment} in {Large} {Language} {Models}: {An} {Explanatory} {Analysis} {Based} on {Hofstede}'s {Cultural} {Dimensions}.
\newblock 2023.
\newblock \doi{10.48550/ARXIV.2309.12342}.
\newblock URL \url{https://arxiv.org/abs/2309.12342}.
\newblock Publisher: arXiv Version Number: 2.

\bibitem[Tao et~al.(2023)Tao, Viberg, Baker, and Kizilcec]{tao_cultural_2023}
Yan Tao, Olga Viberg, Ryan~S. Baker, and René~F. Kizilcec.
\newblock Cultural {Bias} and {Cultural} {Alignment} of {Large} {Language} {Models}.
\newblock November 2023.
\newblock URL \url{https://www.semanticscholar.org/paper/Cultural-Bias-and-Cultural-Alignment-of-Large-Tao-Viberg/5f8bf881c80125452e4a73ad51fdb2c72c65c551}.

\bibitem[Cao et~al.(2023)Cao, Zhou, Lee, Cabello, Chen, and Hershcovich]{cao_assessing_2023}
Yong Cao, Li~Zhou, Seolhwa Lee, Laura Cabello, Min Chen, and Daniel Hershcovich.
\newblock Assessing {Cross}-{Cultural} {Alignment} between {ChatGPT} and {Human} {Societies}: {An} {Empirical} {Study}, March 2023.
\newblock URL \url{http://arxiv.org/abs/2303.17466}.
\newblock arXiv:2303.17466 [cs].

\bibitem[Kirk et~al.(2021)Kirk, Jun, Volpin, Iqbal, Benussi, Dreyer, Shtedritski, and Asano]{kirk_bias_2021}
Hannah~Rose Kirk, Yennie Jun, Filippo Volpin, Haider Iqbal, Elias Benussi, Frederic Dreyer, Aleksandar Shtedritski, and Yuki Asano.
\newblock Bias {Out}-of-the-{Box}: {An} {Empirical} {Analysis} of {Intersectional} {Occupational} {Biases} in {Popular} {Generative} {Language} {Models}.
\newblock In \emph{Advances in {Neural} {Information} {Processing} {Systems}}, volume~34, pages 2611--2624. Curran Associates, Inc., 2021.
\newblock URL \url{https://proceedings.neurips.cc/paper/2021/hash/1531beb762df4029513ebf9295e0d34f-Abstract.html}.

\bibitem[Bender and Koller(2020)]{bender_climbing_2020}
Emily~M. Bender and Alexander Koller.
\newblock Climbing towards {NLU}: {On} {Meaning}, {Form}, and {Understanding} in the {Age} of {Data}.
\newblock In \emph{Proceedings of the 58th {Annual} {Meeting} of the {Association} for {Computational} {Linguistics}}, pages 5185--5198, Online, July 2020. Association for Computational Linguistics.
\newblock \doi{10.18653/v1/2020.acl-main.463}.
\newblock URL \url{https://aclanthology.org/2020.acl-main.463}.

\bibitem[Prashanth et~al.(2024)Prashanth, Deng, O'Brien, S, Khan, Borkar, Choquette-Choo, Fuehne, Biderman, Ke, Lee, and Saphra]{prashanth_recite_2024}
USVSN~Sai Prashanth, Alvin Deng, Kyle O'Brien, Jyothir S, Mohammad~Aflah Khan, Jaydeep Borkar, Christopher~A. Choquette-Choo, Jacob~Ray Fuehne, Stella Biderman, Tracy Ke, Katherine Lee, and Naomi Saphra.
\newblock Recite, {Reconstruct}, {Recollect}: {Memorization} in {LMs} as a {Multifaceted} {Phenomenon}.
\newblock 2024.
\newblock \doi{10.48550/ARXIV.2406.17746}.
\newblock URL \url{https://arxiv.org/abs/2406.17746}.
\newblock Publisher: arXiv Version Number: 1.

\bibitem[Kuhn(1997)]{kuhn_structure_1997}
Thomas~S. Kuhn.
\newblock \emph{The structure of scientific revolutions}, volume 962.
\newblock University of Chicago press Chicago, 1997.
\newblock URL \url{https://www.academia.edu/download/62519294/Thomas_Kuhn_-_The_Structure_of_scientific_revolutions_3rd_ed.20200328-112461-1g7y9qj.pdf}.

\bibitem[Selbst et~al.(2019)Selbst, boyd, Friedler, Venkatasubramanian, and Vertesi]{selbst_fairness_2019}
Andrew~D. Selbst, danah boyd, Sorelle~A. Friedler, Suresh Venkatasubramanian, and Janet Vertesi.
\newblock Fairness and {Abstraction} in {Sociotechnical} {Systems}.
\newblock In \emph{Proceedings of the {Conference} on {Fairness}, {Accountability}, and {Transparency}}, pages 59--68, Atlanta GA USA, January 2019. ACM.
\newblock ISBN 978-1-4503-6125-5.
\newblock \doi{10.1145/3287560.3287598}.
\newblock URL \url{https://dl.acm.org/doi/10.1145/3287560.3287598}.

\bibitem[Burrell(2016)]{burrell_how_2016}
Jenna Burrell.
\newblock How the machine ‘thinks’: {Understanding} opacity in machine learning algorithms.
\newblock \emph{Big Data \& Society}, 3\penalty0 (1):\penalty0 2053951715622512, June 2016.
\newblock ISSN 2053-9517.
\newblock \doi{10.1177/2053951715622512}.
\newblock URL \url{https://ezproxy-prd.bodleian.ox.ac.uk:2246/doi/full/10.1177/2053951715622512}.
\newblock Publisher: SAGE Publications Ltd.

\bibitem[Malik(2020)]{malik_hierarchy_2020}
Momin~M. Malik.
\newblock A {Hierarchy} of {Limitations} in {Machine} {Learning}, February 2020.
\newblock URL \url{http://arxiv.org/abs/2002.05193}.
\newblock arXiv:2002.05193 [cs, econ, math, stat].

\bibitem[Buolamwini and Gebru(2018)]{buolamwini_gender_2018}
Joy Buolamwini and Timnit Gebru.
\newblock Gender {Shades}: {Intersectional} {Accuracy} {Disparities} in {Commercial} {Gender} {Classification}.
\newblock In \emph{Conference on {Fairness}, {Accountability} and {Transparency}}, pages 77--91. PMLR, January 2018.
\newblock URL \url{http://proceedings.mlr.press/v81/buolamwini18a.html}.
\newblock ISSN: 2640-3498.

\bibitem[Birhane et~al.(2022{\natexlab{a}})Birhane, Kalluri, Card, Agnew, Dotan, and Bao]{birhane_values_2022}
Abeba Birhane, Pratyusha Kalluri, Dallas Card, William Agnew, Ravit Dotan, and Michelle Bao.
\newblock The {Values} {Encoded} in {Machine} {Learning} {Research}.
\newblock In \emph{2022 {ACM} {Conference} on {Fairness}, {Accountability}, and {Transparency}}, {FAccT} '22, pages 173--184, New York, NY, USA, June 2022{\natexlab{a}}. Association for Computing Machinery.
\newblock ISBN 978-1-4503-9352-2.
\newblock \doi{10.1145/3531146.3533083}.
\newblock URL \url{https://doi.org/10.1145/3531146.3533083}.

\bibitem[Birhane et~al.(2022{\natexlab{b}})Birhane, Isaac, Prabhakaran, Diaz, Elish, Gabriel, and Mohamed]{birhane_power_2022}
Abeba Birhane, William Isaac, Vinodkumar Prabhakaran, Mark Diaz, Madeleine~Clare Elish, Iason Gabriel, and Shakir Mohamed.
\newblock Power to the {People}? {Opportunities} and {Challenges} for {Participatory} {AI}.
\newblock In \emph{Equity and {Access} in {Algorithms}, {Mechanisms}, and {Optimization}}, {EAAMO} '22, pages 1--8, New York, NY, USA, October 2022{\natexlab{b}}. Association for Computing Machinery.
\newblock ISBN 978-1-4503-9477-2.
\newblock \doi{10.1145/3551624.3555290}.
\newblock URL \url{https://dl.acm.org/doi/10.1145/3551624.3555290}.

\bibitem[Rudin(2019)]{rudin_stop_2019}
Cynthia Rudin.
\newblock Stop explaining black box machine learning models for high stakes decisions and use interpretable models instead.
\newblock \emph{Nature Machine Intelligence}, 1\penalty0 (5):\penalty0 206--215, May 2019.
\newblock ISSN 2522-5839.
\newblock \doi{10.1038/s42256-019-0048-x}.
\newblock URL \url{https://www.nature.com/articles/s42256-019-0048-x}.
\newblock Publisher: Nature Publishing Group.

\bibitem[Pilgrim(2008)]{pilgrim_eugenic_2008}
David Pilgrim.
\newblock The {Eugenic} {Legacy} in {Psychology} and {Psychiatry}.
\newblock \emph{International Journal of Social Psychiatry}, 54\penalty0 (3):\penalty0 272--284, May 2008.
\newblock ISSN 0020-7640.
\newblock \doi{10.1177/0020764008090282}.
\newblock URL \url{https://doi.org/10.1177/0020764008090282}.
\newblock Publisher: SAGE Publications Ltd.

\bibitem[Bender et~al.(2021)Bender, Gebru, McMillan-Major, and Shmitchell]{bender_dangers_2021}
Emily~M. Bender, Timnit Gebru, Angelina McMillan-Major, and Shmargaret Shmitchell.
\newblock On the {Dangers} of {Stochastic} {Parrots}: {Can} {Language} {Models} {Be} {Too} {Big}?
\newblock In \emph{Proceedings of the 2021 {ACM} {Conference} on {Fairness}, {Accountability}, and {Transparency}}, pages 610--623, 2021.

\bibitem[Buzsaki(2019)]{buzsaki2019brain}
Gyorgy Buzsaki.
\newblock \emph{The brain from inside out}.
\newblock Oxford University Press, USA, 2019.

\bibitem[Lee(2011)]{lee_how_2011}
Michael~D. Lee.
\newblock How cognitive modeling can benefit from hierarchical {Bayesian} models.
\newblock \emph{Journal of Mathematical Psychology}, 55\penalty0 (1), February 2011.
\newblock ISSN 0022-2496.
\newblock \doi{10.1016/j.jmp.2010.08.013}.
\newblock URL \url{https://www.sciencedirect.com/science/article/pii/S0022249610001148}.

\bibitem[Zhang et~al.(2023)Zhang, Thakur, Ogundepo, Kamalloo, Alfonso-Hermelo, Li, Liu, Rezagholizadeh, and Lin]{zhang_miracl_2023}
Xinyu Zhang, Nandan Thakur, Odunayo Ogundepo, Ehsan Kamalloo, David Alfonso-Hermelo, Xiaoguang Li, Qun Liu, Mehdi Rezagholizadeh, and Jimmy Lin.
\newblock {MIRACL}: {A} {Multilingual} {Retrieval} {Dataset} {Covering} 18 {Diverse} {Languages}.
\newblock \emph{Transactions of the Association for Computational Linguistics}, 11:\penalty0 1114--1131, September 2023.
\newblock ISSN 2307-387X.
\newblock \doi{10.1162/tacl_a_00595}.
\newblock URL \url{https://doi.org/10.1162/tacl_a_00595}.

\bibitem[Hu et~al.(2020)Hu, Ruder, Siddhant, Neubig, Firat, and Johnson]{hu_xtreme_2020}
Junjie Hu, Sebastian Ruder, Aditya Siddhant, Graham Neubig, Orhan Firat, and Melvin Johnson.
\newblock {XTREME}: {A} {Massively} {Multilingual} {Multi}-task {Benchmark} for {Evaluating} {Cross}-lingual {Generalisation}.
\newblock In \emph{Proceedings of the 37th {International} {Conference} on {Machine} {Learning}}, pages 4411--4421. PMLR, November 2020.
\newblock URL \url{https://proceedings.mlr.press/v119/hu20b.html}.
\newblock ISSN: 2640-3498.

\bibitem[Hermann et~al.(2015)Hermann, Kocisky, Grefenstette, Espeholt, Kay, Suleyman, and Blunsom]{hermann_teaching_2015}
Karl~Moritz Hermann, Tomas Kocisky, Edward Grefenstette, Lasse Espeholt, Will Kay, Mustafa Suleyman, and Phil Blunsom.
\newblock Teaching {Machines} to {Read} and {Comprehend}.
\newblock In \emph{Advances in {Neural} {Information} {Processing} {Systems}}, volume~28. Curran Associates, Inc., 2015.
\newblock URL \url{https://proceedings.neurips.cc/paper_files/paper/2015/hash/afdec7005cc9f14302cd0474fd0f3c96-Abstract.html}.

\bibitem[Rajpurkar et~al.(2016)Rajpurkar, Zhang, Lopyrev, and Liang]{rajpurkar_squad_2016}
Pranav Rajpurkar, Jian Zhang, Konstantin Lopyrev, and Percy Liang.
\newblock {SQuAD}: 100,000+ {Questions} for {Machine} {Comprehension} of {Text}, October 2016.
\newblock URL \url{http://arxiv.org/abs/1606.05250}.
\newblock arXiv:1606.05250 [cs].

\bibitem[Nielsen(2023)]{nielsen_scandeval_2023}
Dan~Saattrup Nielsen.
\newblock {ScandEval}: {A} {Benchmark} for {Scandinavian} {Natural} {Language} {Processing}, April 2023.
\newblock URL \url{http://arxiv.org/abs/2304.00906}.
\newblock arXiv:2304.00906 [cs].

\bibitem[Nielsen et~al.(2024)Nielsen, Enevoldsen, and Schneider-Kamp]{nielsen_encoder_2024}
Dan~Saattrup Nielsen, Kenneth Enevoldsen, and Peter Schneider-Kamp.
\newblock Encoder vs {Decoder}: {Comparative} {Analysis} of {Encoder} and {Decoder} {Language} {Models} on {Multilingual} {NLU} {Tasks}, June 2024.
\newblock URL \url{http://arxiv.org/abs/2406.13469}.
\newblock arXiv:2406.13469 [cs].

\bibitem[OpenAI(2023)]{openai_gpt-4_2023}
OpenAI.
\newblock {GPT}-4 {Technical} {Report}.
\newblock Technical report, OpenAI, March 2023.
\newblock URL \url{https://cdn.openai.com/papers/gpt-4.pdf}.

\bibitem[Touvron et~al.(2023)Touvron, Lavril, Izacard, Martinet, Lachaux, Lacroix, Rozière, Goyal, Hambro, Azhar, Rodriguez, Joulin, Grave, and Lample]{touvron_llama_2023}
Hugo Touvron, Thibaut Lavril, Gautier Izacard, Xavier Martinet, Marie-Anne Lachaux, Timothée Lacroix, Baptiste Rozière, Naman Goyal, Eric Hambro, Faisal Azhar, Aurelien Rodriguez, Armand Joulin, Edouard Grave, and Guillaume Lample.
\newblock {LLaMA}: {Open} and {Efficient} {Foundation} {Language} {Models}, February 2023.
\newblock URL \url{http://arxiv.org/abs/2302.13971}.
\newblock arXiv:2302.13971 [cs].

\bibitem[Jiang et~al.(2023)Jiang, Sablayrolles, Mensch, Bamford, Chaplot, Casas, Bressand, Lengyel, Lample, Saulnier, Lavaud, Lachaux, Stock, Scao, Lavril, Wang, Lacroix, and Sayed]{jiang_mistral_2023}
Albert~Q. Jiang, Alexandre Sablayrolles, Arthur Mensch, Chris Bamford, Devendra~Singh Chaplot, Diego de~las Casas, Florian Bressand, Gianna Lengyel, Guillaume Lample, Lucile Saulnier, Lélio~Renard Lavaud, Marie-Anne Lachaux, Pierre Stock, Teven~Le Scao, Thibaut Lavril, Thomas Wang, Timothée Lacroix, and William~El Sayed.
\newblock Mistral {7B}, October 2023.
\newblock URL \url{http://arxiv.org/abs/2310.06825}.
\newblock arXiv:2310.06825 [cs].

\bibitem[Igolkina and Meshcheryakov(2020)]{igolkina_semopy_2020}
Anna~A. Igolkina and Georgy Meshcheryakov.
\newblock semopy: {A} {Python} {Package} for {Structural} {Equation} {Modeling}.
\newblock \emph{Structural Equation Modeling: A Multidisciplinary Journal}, 27\penalty0 (6):\penalty0 952--963, November 2020.
\newblock ISSN 1070-5511.
\newblock \doi{10.1080/10705511.2019.1704289}.
\newblock URL \url{https://doi.org/10.1080/10705511.2019.1704289}.
\newblock Publisher: Routledge \_eprint: https://doi.org/10.1080/10705511.2019.1704289.

\bibitem[Hu and Bentler(1999)]{hu_cutoff_1999}
Li‐tze Hu and Peter~M. Bentler.
\newblock Cutoff criteria for fit indexes in covariance structure analysis: {Conventional} criteria versus new alternatives.
\newblock \emph{Structural Equation Modeling: A Multidisciplinary Journal}, 6\penalty0 (1):\penalty0 1--55, January 1999.
\newblock ISSN 1070-5511, 1532-8007.
\newblock \doi{10.1080/10705519909540118}.
\newblock URL \url{http://www.tandfonline.com/doi/abs/10.1080/10705519909540118}.

\bibitem[Gigerenzer(2004)]{gigerenzer_mindless_2004}
Gerd Gigerenzer.
\newblock Mindless statistics.
\newblock \emph{The Journal of Socio-Economics}, 33\penalty0 (5):\penalty0 587--606, November 2004.
\newblock ISSN 1053-5357.
\newblock \doi{10.1016/j.socec.2004.09.033}.
\newblock URL \url{https://www.sciencedirect.com/science/article/pii/S1053535704000927}.

\bibitem[Sullivan and Feinn(2012)]{sullivan_using_2012}
Gail~M. Sullivan and Richard Feinn.
\newblock Using {Effect} {Size}—or {Why} the {P} {Value} {Is} {Not} {Enough}.
\newblock \emph{Journal of Graduate Medical Education}, 4\penalty0 (3):\penalty0 279--282, September 2012.
\newblock ISSN 1949-8349.
\newblock \doi{10.4300/JGME-D-12-00156.1}.
\newblock URL \url{https://www.ncbi.nlm.nih.gov/pmc/articles/PMC3444174/}.

\bibitem[Baayen et~al.(2008)Baayen, Davidson, and Bates]{baayen_mixed-effects_2008}
R.~Harald Baayen, Douglas~J. Davidson, and Douglas~M. Bates.
\newblock Mixed-effects modeling with crossed random effects for subjects and items.
\newblock \emph{Journal of memory and language}, 59\penalty0 (4):\penalty0 390--412, 2008.
\newblock Publisher: Elsevier.

\bibitem[Meshcheryakov et~al.(2021)Meshcheryakov, Igolkina, and Samsonova]{meshcheryakov_semopy_2021}
Georgy Meshcheryakov, Anna~A. Igolkina, and Maria~G. Samsonova.
\newblock semopy 2: {A} {Structural} {Equation} {Modeling} {Package} with {Random} {Effects} in {Python}, June 2021.
\newblock URL \url{https://arxiv.org/abs/2106.01140v3}.

\bibitem[Srivastava et~al.(2022)Srivastava, Rastogi, Rao, Shoeb, Abid, Fisch, Brown, Santoro, Gupta, Garriga-Alonso, Kluska, Lewkowycz, Agarwal, Power, Ray, Warstadt, Kocurek, Safaya, Tazarv, Xiang, Parrish, Nie, Hussain, Askell, Dsouza, Slone, Rahane, Iyer, Andreassen, Madotto, Santilli, Stuhlmüller, Dai, La, Lampinen, Zou, Jiang, Chen, Vuong, Gupta, Gottardi, Norelli, Venkatesh, Gholamidavoodi, Tabassum, Menezes, Kirubarajan, Mullokandov, Sabharwal, Herrick, Efrat, Erdem, Karakaş, Roberts, Loe, Zoph, Bojanowski, Özyurt, Hedayatnia, Neyshabur, Inden, Stein, Ekmekci, Lin, Howald, Diao, Dour, Stinson, Argueta, Ramírez, Singh, Rathkopf, Meng, Baral, Wu, Callison-Burch, Waites, Voigt, Manning, Potts, Ramirez, Rivera, Siro, Raffel, Ashcraft, Garbacea, Sileo, Garrette, Hendrycks, Kilman, Roth, Freeman, Khashabi, Levy, González, Perszyk, Hernandez, Chen, Ippolito, Gilboa, Dohan, Drakard, Jurgens, Datta, Ganguli, Emelin, Kleyko, Yuret, Chen, Tam, Hupkes, Misra, Buzan, Mollo, Yang, Lee, Shutova, Cubuk, Segal,
  Hagerman, Barnes, Donoway, Pavlick, Rodola, Lam, Chu, Tang, Erdem, Chang, Chi, Dyer, Jerzak, Kim, Manyasi, Zheltonozhskii, Xia, Siar, Martínez-Plumed, Happé, Chollet, Rong, Mishra, Winata, de~Melo, Kruszewski, Parascandolo, Mariani, Wang, Jaimovitch-López, Betz, Gur-Ari, Galijasevic, Kim, Rashkin, Hajishirzi, Mehta, Bogar, Shevlin, Schütze, Yakura, Zhang, Wong, Ng, Noble, Jumelet, Geissinger, Kernion, Hilton, Lee, Fisac, Simon, Koppel, Zheng, Zou, Kocoń, Thompson, Kaplan, Radom, Sohl-Dickstein, Phang, Wei, Yosinski, Novikova, Bosscher, Marsh, Kim, Taal, Engel, Alabi, Xu, Song, Tang, Waweru, Burden, Miller, Balis, Berant, Frohberg, Rozen, Hernandez-Orallo, Boudeman, Jones, Tenenbaum, Rule, Chua, Kanclerz, Livescu, Krauth, Gopalakrishnan, Ignatyeva, Markert, Dhole, Gimpel, Omondi, Mathewson, Chiafullo, Shkaruta, Shridhar, McDonell, Richardson, Reynolds, Gao, Zhang, Dugan, Qin, Contreras-Ochando, Morency, Moschella, Lam, Noble, Schmidt, He, Colón, Metz, Şenel, Bosma, Sap, ter Hoeve, Farooqi, Faruqui,
  Mazeika, Baturan, Marelli, Maru, Quintana, Tolkiehn, Giulianelli, Lewis, Potthast, Leavitt, Hagen, Schubert, Baitemirova, Arnaud, McElrath, Yee, Cohen, Gu, Ivanitskiy, Starritt, Strube, Swędrowski, Bevilacqua, Yasunaga, Kale, Cain, Xu, Suzgun, Tiwari, Bansal, Aminnaseri, Geva, Gheini, T, Peng, Chi, Lee, Krakover, Cameron, Roberts, Doiron, Nangia, Deckers, Muennighoff, Keskar, Iyer, Constant, Fiedel, Wen, Zhang, Agha, Elbaghdadi, Levy, Evans, Casares, Doshi, Fung, Liang, Vicol, Alipoormolabashi, Liao, Liang, Chang, Eckersley, Htut, Hwang, Miłkowski, Patil, Pezeshkpour, Oli, Mei, Lyu, Chen, Banjade, Rudolph, Gabriel, Habacker, Delgado, Millière, Garg, Barnes, Saurous, Arakawa, Raymaekers, Frank, Sikand, Novak, Sitelew, LeBras, Liu, Jacobs, Zhang, Salakhutdinov, Chi, Lee, Stovall, Teehan, Yang, Singh, Mohammad, Anand, Dillavou, Shleifer, Wiseman, Gruetter, Bowman, Schoenholz, Han, Kwatra, Rous, Ghazarian, Ghosh, Casey, Bischoff, Gehrmann, Schuster, Sadeghi, Hamdan, Zhou, Srivastava, Shi, Singh, Asaadi, Gu,
  Pachchigar, Toshniwal, Upadhyay, Shyamolima, Debnath, Shakeri, Thormeyer, Melzi, Reddy, Makini, Lee, Torene, Hatwar, Dehaene, Divic, Ermon, Biderman, Lin, Prasad, Piantadosi, Shieber, Misherghi, Kiritchenko, Mishra, Linzen, Schuster, Li, Yu, Ali, Hashimoto, Wu, Desbordes, Rothschild, Phan, Wang, Nkinyili, Schick, Kornev, Telleen-Lawton, Tunduny, Gerstenberg, Chang, Neeraj, Khot, Shultz, Shaham, Misra, Demberg, Nyamai, Raunak, Ramasesh, Prabhu, Padmakumar, Srikumar, Fedus, Saunders, Zhang, Vossen, Ren, Tong, Zhao, Wu, Shen, Yaghoobzadeh, Lakretz, Song, Bahri, Choi, Yang, Hao, Chen, Belinkov, Hou, Hou, Bai, Seid, Zhao, Wang, Wang, Wang, and Wu]{srivastava_beyond_2022}
Aarohi Srivastava, Abhinav Rastogi, Abhishek Rao, Abu Awal~Md Shoeb, Abubakar Abid, Adam Fisch, Adam~R. Brown, Adam Santoro, Aditya Gupta, Adrià Garriga-Alonso, Agnieszka Kluska, Aitor Lewkowycz, Akshat Agarwal, Alethea Power, Alex Ray, Alex Warstadt, Alexander~W. Kocurek, Ali Safaya, Ali Tazarv, Alice Xiang, Alicia Parrish, Allen Nie, Aman Hussain, Amanda Askell, Amanda Dsouza, Ambrose Slone, Ameet Rahane, Anantharaman~S. Iyer, Anders Andreassen, Andrea Madotto, Andrea Santilli, Andreas Stuhlmüller, Andrew Dai, Andrew La, Andrew Lampinen, Andy Zou, Angela Jiang, Angelica Chen, Anh Vuong, Animesh Gupta, Anna Gottardi, Antonio Norelli, Anu Venkatesh, Arash Gholamidavoodi, Arfa Tabassum, Arul Menezes, Arun Kirubarajan, Asher Mullokandov, Ashish Sabharwal, Austin Herrick, Avia Efrat, Aykut Erdem, Ayla Karakaş, B.~Ryan Roberts, Bao~Sheng Loe, Barret Zoph, Bartłomiej Bojanowski, Batuhan Özyurt, Behnam Hedayatnia, Behnam Neyshabur, Benjamin Inden, Benno Stein, Berk Ekmekci, Bill~Yuchen Lin, Blake Howald,
  Cameron Diao, Cameron Dour, Catherine Stinson, Cedrick Argueta, César~Ferri Ramírez, Chandan Singh, Charles Rathkopf, Chenlin Meng, Chitta Baral, Chiyu Wu, Chris Callison-Burch, Chris Waites, Christian Voigt, Christopher~D. Manning, Christopher Potts, Cindy Ramirez, Clara~E. Rivera, Clemencia Siro, Colin Raffel, Courtney Ashcraft, Cristina Garbacea, Damien Sileo, Dan Garrette, Dan Hendrycks, Dan Kilman, Dan Roth, Daniel Freeman, Daniel Khashabi, Daniel Levy, Daniel~Moseguí González, Danielle Perszyk, Danny Hernandez, Danqi Chen, Daphne Ippolito, Dar Gilboa, David Dohan, David Drakard, David Jurgens, Debajyoti Datta, Deep Ganguli, Denis Emelin, Denis Kleyko, Deniz Yuret, Derek Chen, Derek Tam, Dieuwke Hupkes, Diganta Misra, Dilyar Buzan, Dimitri~Coelho Mollo, Diyi Yang, Dong-Ho Lee, Ekaterina Shutova, Ekin~Dogus Cubuk, Elad Segal, Eleanor Hagerman, Elizabeth Barnes, Elizabeth Donoway, Ellie Pavlick, Emanuele Rodola, Emma Lam, Eric Chu, Eric Tang, Erkut Erdem, Ernie Chang, Ethan~A. Chi, Ethan Dyer, Ethan
  Jerzak, Ethan Kim, Eunice~Engefu Manyasi, Evgenii Zheltonozhskii, Fanyue Xia, Fatemeh Siar, Fernando Martínez-Plumed, Francesca Happé, Francois Chollet, Frieda Rong, Gaurav Mishra, Genta~Indra Winata, Gerard de~Melo, Germán Kruszewski, Giambattista Parascandolo, Giorgio Mariani, Gloria Wang, Gonzalo Jaimovitch-López, Gregor Betz, Guy Gur-Ari, Hana Galijasevic, Hannah Kim, Hannah Rashkin, Hannaneh Hajishirzi, Harsh Mehta, Hayden Bogar, Henry Shevlin, Hinrich Schütze, Hiromu Yakura, Hongming Zhang, Hugh~Mee Wong, Ian Ng, Isaac Noble, Jaap Jumelet, Jack Geissinger, Jackson Kernion, Jacob Hilton, Jaehoon Lee, Jaime~Fernández Fisac, James~B. Simon, James Koppel, James Zheng, James Zou, Jan Kocoń, Jana Thompson, Jared Kaplan, Jarema Radom, Jascha Sohl-Dickstein, Jason Phang, Jason Wei, Jason Yosinski, Jekaterina Novikova, Jelle Bosscher, Jennifer Marsh, Jeremy Kim, Jeroen Taal, Jesse Engel, Jesujoba Alabi, Jiacheng Xu, Jiaming Song, Jillian Tang, Joan Waweru, John Burden, John Miller, John~U. Balis,
  Jonathan Berant, Jörg Frohberg, Jos Rozen, Jose Hernandez-Orallo, Joseph Boudeman, Joseph Jones, Joshua~B. Tenenbaum, Joshua~S. Rule, Joyce Chua, Kamil Kanclerz, Karen Livescu, Karl Krauth, Karthik Gopalakrishnan, Katerina Ignatyeva, Katja Markert, Kaustubh~D. Dhole, Kevin Gimpel, Kevin Omondi, Kory Mathewson, Kristen Chiafullo, Ksenia Shkaruta, Kumar Shridhar, Kyle McDonell, Kyle Richardson, Laria Reynolds, Leo Gao, Li~Zhang, Liam Dugan, Lianhui Qin, Lidia Contreras-Ochando, Louis-Philippe Morency, Luca Moschella, Lucas Lam, Lucy Noble, Ludwig Schmidt, Luheng He, Luis~Oliveros Colón, Luke Metz, Lütfi~Kerem Şenel, Maarten Bosma, Maarten Sap, Maartje ter Hoeve, Maheen Farooqi, Manaal Faruqui, Mantas Mazeika, Marco Baturan, Marco Marelli, Marco Maru, Maria Jose~Ramírez Quintana, Marie Tolkiehn, Mario Giulianelli, Martha Lewis, Martin Potthast, Matthew~L. Leavitt, Matthias Hagen, Mátyás Schubert, Medina~Orduna Baitemirova, Melody Arnaud, Melvin McElrath, Michael~A. Yee, Michael Cohen, Michael Gu,
  Michael Ivanitskiy, Michael Starritt, Michael Strube, Michał Swędrowski, Michele Bevilacqua, Michihiro Yasunaga, Mihir Kale, Mike Cain, Mimee Xu, Mirac Suzgun, Mo~Tiwari, Mohit Bansal, Moin Aminnaseri, Mor Geva, Mozhdeh Gheini, Mukund~Varma T, Nanyun Peng, Nathan Chi, Nayeon Lee, Neta Gur-Ari Krakover, Nicholas Cameron, Nicholas Roberts, Nick Doiron, Nikita Nangia, Niklas Deckers, Niklas Muennighoff, Nitish~Shirish Keskar, Niveditha~S. Iyer, Noah Constant, Noah Fiedel, Nuan Wen, Oliver Zhang, Omar Agha, Omar Elbaghdadi, Omer Levy, Owain Evans, Pablo Antonio~Moreno Casares, Parth Doshi, Pascale Fung, Paul~Pu Liang, Paul Vicol, Pegah Alipoormolabashi, Peiyuan Liao, Percy Liang, Peter Chang, Peter Eckersley, Phu~Mon Htut, Pinyu Hwang, Piotr Miłkowski, Piyush Patil, Pouya Pezeshkpour, Priti Oli, Qiaozhu Mei, Qing Lyu, Qinlang Chen, Rabin Banjade, Rachel~Etta Rudolph, Raefer Gabriel, Rahel Habacker, Ramón~Risco Delgado, Raphaël Millière, Rhythm Garg, Richard Barnes, Rif~A. Saurous, Riku Arakawa, Robbe
  Raymaekers, Robert Frank, Rohan Sikand, Roman Novak, Roman Sitelew, Ronan LeBras, Rosanne Liu, Rowan Jacobs, Rui Zhang, Ruslan Salakhutdinov, Ryan Chi, Ryan Lee, Ryan Stovall, Ryan Teehan, Rylan Yang, Sahib Singh, Saif~M. Mohammad, Sajant Anand, Sam Dillavou, Sam Shleifer, Sam Wiseman, Samuel Gruetter, Samuel~R. Bowman, Samuel~S. Schoenholz, Sanghyun Han, Sanjeev Kwatra, Sarah~A. Rous, Sarik Ghazarian, Sayan Ghosh, Sean Casey, Sebastian Bischoff, Sebastian Gehrmann, Sebastian Schuster, Sepideh Sadeghi, Shadi Hamdan, Sharon Zhou, Shashank Srivastava, Sherry Shi, Shikhar Singh, Shima Asaadi, Shixiang~Shane Gu, Shubh Pachchigar, Shubham Toshniwal, Shyam Upadhyay, Shyamolima, Debnath, Siamak Shakeri, Simon Thormeyer, Simone Melzi, Siva Reddy, Sneha~Priscilla Makini, Soo-Hwan Lee, Spencer Torene, Sriharsha Hatwar, Stanislas Dehaene, Stefan Divic, Stefano Ermon, Stella Biderman, Stephanie Lin, Stephen Prasad, Steven~T. Piantadosi, Stuart~M. Shieber, Summer Misherghi, Svetlana Kiritchenko, Swaroop Mishra, Tal
  Linzen, Tal Schuster, Tao Li, Tao Yu, Tariq Ali, Tatsu Hashimoto, Te-Lin Wu, Théo Desbordes, Theodore Rothschild, Thomas Phan, Tianle Wang, Tiberius Nkinyili, Timo Schick, Timofei Kornev, Timothy Telleen-Lawton, Titus Tunduny, Tobias Gerstenberg, Trenton Chang, Trishala Neeraj, Tushar Khot, Tyler Shultz, Uri Shaham, Vedant Misra, Vera Demberg, Victoria Nyamai, Vikas Raunak, Vinay Ramasesh, Vinay~Uday Prabhu, Vishakh Padmakumar, Vivek Srikumar, William Fedus, William Saunders, William Zhang, Wout Vossen, Xiang Ren, Xiaoyu Tong, Xinran Zhao, Xinyi Wu, Xudong Shen, Yadollah Yaghoobzadeh, Yair Lakretz, Yangqiu Song, Yasaman Bahri, Yejin Choi, Yichi Yang, Yiding Hao, Yifu Chen, Yonatan Belinkov, Yu~Hou, Yufang Hou, Yuntao Bai, Zachary Seid, Zhuoye Zhao, Zijian Wang, Zijie~J. Wang, Zirui Wang, and Ziyi Wu.
\newblock Beyond the {Imitation} {Game}: {Quantifying} and extrapolating the capabilities of language models, June 2022.
\newblock URL \url{http://arxiv.org/abs/2206.04615}.
\newblock arXiv:2206.04615 [cs, stat].

\bibitem[Warstadt et~al.(2019)Warstadt, Singh, and Bowman]{warstadt_neural_2019}
Alex Warstadt, Amanpreet Singh, and Samuel~R. Bowman.
\newblock Neural {Network} {Acceptability} {Judgments}.
\newblock \emph{Transactions of the Association for Computational Linguistics}, 7:\penalty0 625--641, September 2019.
\newblock ISSN 2307-387X.
\newblock \doi{10.1162/tacl_a_00290}.
\newblock URL \url{https://doi.org/10.1162/tacl_a_00290}.

\bibitem[Sap et~al.(2019)Sap, Rashkin, Chen, Le~Bras, and Choi]{sap_social_2019}
Maarten Sap, Hannah Rashkin, Derek Chen, Ronan Le~Bras, and Yejin Choi.
\newblock Social {IQa}: {Commonsense} {Reasoning} about {Social} {Interactions}.
\newblock In Kentaro Inui, Jing Jiang, Vincent Ng, and Xiaojun Wan, editors, \emph{Proceedings of the 2019 {Conference} on {Empirical} {Methods} in {Natural} {Language} {Processing} and the 9th {International} {Joint} {Conference} on {Natural} {Language} {Processing} ({EMNLP}-{IJCNLP})}, pages 4463--4473, Hong Kong, China, November 2019. Association for Computational Linguistics.
\newblock \doi{10.18653/v1/D19-1454}.
\newblock URL \url{https://aclanthology.org/D19-1454}.

\bibitem[Askell et~al.(2021)Askell, Bai, Chen, Drain, Ganguli, Henighan, Jones, Joseph, Mann, DasSarma, Elhage, Hatfield-Dodds, Hernandez, Kernion, Ndousse, Olsson, Amodei, Brown, Clark, McCandlish, Olah, and Kaplan]{askell_general_2021}
Amanda Askell, Yuntao Bai, Anna Chen, Dawn Drain, Deep Ganguli, Tom Henighan, Andy Jones, Nicholas Joseph, Ben Mann, Nova DasSarma, Nelson Elhage, Zac Hatfield-Dodds, Danny Hernandez, Jackson Kernion, Kamal Ndousse, Catherine Olsson, Dario Amodei, Tom Brown, Jack Clark, Sam McCandlish, Chris Olah, and Jared Kaplan.
\newblock A {General} {Language} {Assistant} as a {Laboratory} for {Alignment}, December 2021.
\newblock URL \url{http://arxiv.org/abs/2112.00861}.
\newblock arXiv:2112.00861 [cs].

\end{thebibliography}

\appendix
\section{Estimating Cross-cultural Language Ability: A Simplified Example} \label{app:simplified-sem}

\begin{figure}[hbtp]
    \centering
    \includegraphics[width=0.85\linewidth]{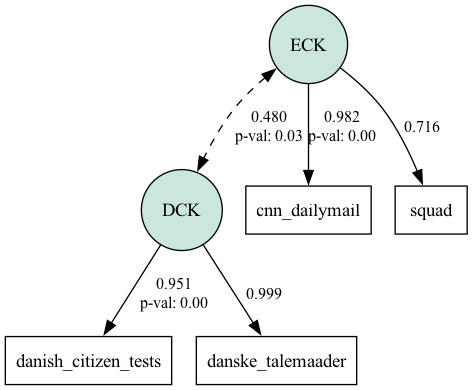}
    \caption{Caption}
    \label{fig:app:toy_model}
\end{figure}

This appendix provides a detailed illustration of building a cognitive model using Structural Equation Modelling (SEM), focusing on a toy example of knowledge transfer in Large Language Models (LLMs). While not intended as a rigorous empirical analysis \citep[see, e.g.,][]{zhang_miracl_2023,hu_xtreme_2020}, it offers insights into how such a model could be constructed. Specifically, this example demonstrates how cognitive models, operationalised as SEMs, can expose assumptions and answer questions about LLMs, supporting the overarching framework presented in the main text. All code for reproducing the analysis can be found in this repository\footnote{https://anonymous.4open.science/r/llatent-psych-EB2D/}.

The hypothesis we set out to test is the following:

\begin{itemize}
    \item \textbf{RQ1}: Does an underlying construct connect Danish Cultural Knowledge and English Cultural Knowledge in LLMs?
\end{itemize}

To test this, we construct the SEM shown in Fig. \ref{fig:app:toy_model}. The model is quite simple: it has two latent variables for Danish Cultural Knowledge ($DCK$) and English Cultural Knowledge ($ECK$) with a proposed co-variance between them (dashed line). 

Each latent variable has two associated benchmarks (square boxes). For Danish, these are \texttt{danish\_citizen\_tests} and \texttt{danske\_talemaader} (see appendix \ref{app:benchmark} for descriptions). For English, we use the CNN/Dailymail dataset, a news summarization task \citep{hermann_teaching_2015}, and SQuAD, a canonical question-answering dataset \citep{rajpurkar_squad_2016}. 

\paragraph{Data}
The datasets were chosen primarily for convenience, as they are all part of ScandEval, a multilingual benchmark for comparing LLMs across Germanic languages including the Scandinavian languages as well as Dutch, English, and German \citep{nielsen_scandeval_2023,nielsen_encoder_2024}. Specifically, we use the Germanic Natural Language Generation-subset\footnote{https://scandeval.com/germanic-nlg/}, which provides responses for both Danish and English across a wide range of LLMs. While these datasets serve our purpose in this appendix, a more rigorous study would require carefully selecting datasets more accurately representing cultural knowledge across languages.

The specific LLMs tested in ScandEval include a mix of commercial closed-source models, like OpenAI's GPT-series \citep{openai_gpt-4_2023}, as well as open-source base models like LLAMA \citep{touvron_llama_2023} and Mistral \citep{jiang_mistral_2023}, along with various fine-tuned versions of these models available on Hugging Face\footnote{https://huggingface.co/models}. In total, the dataset contains responses from 45 LLMs.

\paragraph{Modelling and Hypothesis Testing}
To conduct the analysis, we use the software \texttt{semopy}, a Python framework for optimizing SEMs \citep{igolkina_semopy_2020}. To gauge model fit, we rely on the widely adopted criteria by \citet{hu_cutoff_1999}, specifically a Comparative Fit Index (CFI) of above 0.95 and a Root Mean Squared Error of Approximation (RMSEA) of less than 0.06 \citep{suhr_basics_2006}. However, we caution against the dogmatic use of test statistics - this is not a practice we want to import from social science \citep[see, e.g.,][]{gigerenzer_mindless_2004}.

To test the hypothesis, we assess the value of the correlation coefficient between $ECK$ and $DCK$. If it's significant, positive and non-trivial, we affirm our research question \citep{sullivan_using_2012}. 

It's worth noting that while the computational aspects of SEM are well-established and efficient, the main constraint in scaling this approach lies in generating the necessary benchmark data. Evaluating LLMs across multiple tasks and languages requires significant computational resources, which should be considered when designing larger-scale studies.

The code was run locally on MacBook Air with an M3 running Sonoma 14.6.1. See \href{https://anonymous.4open.science/r/llatent-psych-EB2D}{the repository} for cross-platform setup instructions.

\paragraph{Results}
The main outcomes of the analysis can be seen in Fig. \ref{fig:app:toy_model}. The analysis reveals a meaningful and statistically significant positive correlation between Danish Cultural Knowledge (DCK) and English Cultural Knowledge (ECK) ($\rho = 0.48$, $p = 0.03$). All tasks show significant loadings on their respective latent variables. The model demonstrates an excellent fit, with an RMSEA near zero and a CFI above 1.

Thus, the model provides strong evidence of an underlying factor explaining Danish Cultural Knowledge and English Cultural Knowledge - albeit on a limited dataset. However, as noted in our limitations, this latent variable might represent general language ability rather than specific cultural knowledge, underscoring the importance of careful construct validity in the cognitive modelling of LLMs.

\paragraph{Learnings from the Analysis}
While this example is obviously simplified, it highlights important considerations for conducting cognitive modelling of LLMs in practice:

\begin{enumerate}
    \item \textbf{Concept Validity is (still) key}: SEMs provide an elegant way to test relationships between latent variables. However, whether these variables represent valid constructs in the real world requires careful theoretical considerations \citep{jacobs_measurement_2021,hughes_psychometric_2018}. For instance, in this example, `Cultural Knowledge' might just as well capture `Language Ability' \citep[see][]{chollet_measure_2019} or some other factor.
    \item \textbf{Sufficient (Statistical) Power}: Testing complex, realistic models requires larger datasets. This is challenging when each LLM run only provides a single data point (as in this example). This limitation can be addressed by employing repeated measures models \citep{baayen_mixed-effects_2008}, where each LLM is sampled multiple times. Such analysis is supported by \texttt{semopy} \citep{meshcheryakov_semopy_2021}.
    \item \textbf{Don't start from scratch}: SEM is not a new tool. Thousands of papers from economics to biology use it. Many solutions to technical and theoretical problems have likely been addressed by statisticians, economists, or biologists in other fields.
\end{enumerate}

While this appendix presents a simplified example, the main paper explores how this approach can be expanded into a comprehensive framework for evaluating LLM capabilities across languages and cultures. The insights gained from this toy model inform the broader discussion on developing robust evaluation methods for Generative AI.

\section{Benchmark Descriptions} \label{app:benchmark}

Table \ref{tab:benchmarks} provides an overview of the benchmarks used in our Structural Equation Model for Cross-lingual Alignment Transfer (Fig. \ref{fig:main-concept-model}.). The English benchmarks are taken from BIG-bench \citep{srivastava_beyond_2022}, which is released under an Apache 2.0 license. The Danish tasks are from ScandEval \citep{nielsen_scandeval_2023}, which is released under an MIT license.

\begin{table}[hbpt]
  \caption{Benchmark descriptions and latent factor loadings}
  \label{tab:benchmarks}
  \centering
  \resizebox{\textwidth}{!}{%
  \begin{tabular}{p{1.5cm}p{4cm}p{3.5cm}p{3.5cm}}
    \toprule
    Benchmark & Description & Concept Loading & Reference \\
    \midrule
    \multicolumn{4}{l}{\textit{English benchmarks}} \\
    \cmidrule(r){1-4}
    CoLA & Binary single-sentence classification task for linguistic acceptability & G\textsubscript{eng} & \citet{warstadt_neural_2019} \\
    SocialIQA & Multiple-choice QA task on social situations and relationships & G\textsubscript{eng}, CK\textsubscript{eng} & \citet{sap_social_2019} \\
    English Proverbs & Tests contextual knowledge of English proverbs & G\textsubscript{eng}, CK\textsubscript{eng} & \citet{srivastava_beyond_2022} \\
    HHH & Evaluates LLMs on being harmless, helpful, and honest in various scenarios & G\textsubscript{eng}, CK\textsubscript{eng}, HHH\textsubscript{eng} & \citet{askell_general_2021} \\
    \midrule
    \multicolumn{4}{l}{\textit{Danish benchmarks}} \\
    \cmidrule(r){1-4}
    ScaLA & Adapted NLP tasks for Danish language understanding and generation & G\textsubscript{dan} & \citet{nielsen_scandeval_2023} \\
    Talemaader & Task focused on understanding and using Danish idioms and expressions & G\textsubscript{dan}, CK\textsubscript{dan} & \citet{nielsen_encoder_2024} \\
    Integrations-proeven & Test assessing Danish Cultural Knowledge normally used for immigration & G\textsubscript{dan}, CK\textsubscript{dan} & \citet{nielsen_encoder_2024} \\
    Danish HHH* & Proposed benchmark to assess harmlessness, helpfulness, and honesty in Danish contexts & G\textsubscript{dan}, CK\textsubscript{dan}, HHH\textsubscript{dan} & N/A \\
    \bottomrule
    \multicolumn{4}{l}{\small *Hypothetical or proposed benchmark} \\
    \multicolumn{4}{l}{\small G: Language Ability, CK: Cultural Knowledge, HHH: Alignment (Harmless, Helpful, Honest)} \\
  \end{tabular}
  }
\end{table}

\end{document}